\documentclass[onecolumn]{ceurart}

\usepackage{graphicx}
\usepackage{amsmath}
\usepackage{booktabs}
\usepackage{caption}
\usepackage{subcaption}
\usepackage{multirow}
\usepackage{float}

\providecommand{\Description}[1]{}

\usepackage{listings}
\usepackage{graphicx}
\usepackage{subcaption}
\usepackage{placeins}  

\copyrightyear{2025}
\copyrightclause{Copyright for this paper by its authors.
  Use permitted under Creative Commons License Attribution 4.0
  International (CC BY 4.0).}

\conference{AIP2025: Second Workshop on AI in Production, 
  2025}

\begin{document}

\title{Fusion-Based Neural Generalization for Predicting Temperature Fields in Industrial PET Preform Heating}

\author[1,2]{Ahmad Alsheikh}[email=ahmad.alsheikh@krones.com]
\author[2]{Andreas Fischer}[email=andreas.fischer@th-deg.de]

\address[1]{KRONES AG, Böhmerwaldstr. 5, 93073 Neutraubling, Germany}
\address[2]{Deggendorf Institute of Technology, Dieter-Görlitz-Platz 1, 94469 Deggendorf, Germany}

\begin{abstract}
Accurate and efficient temperature prediction is critical for optimizing the preheating process of PET preforms in industrial microwave systems prior to blow molding. We propose a novel deep learning framework for generalized temperature prediction. Unlike traditional models that require extensive retraining for each material or design variation, our method introduces a data-efficient neural architecture that leverages transfer learning and model fusion to generalize across unseen scenarios. By pretraining specialized neural regressor on distinct conditions such as recycled PET heat capacities or varying preform geometries and integrating their representations into a unified global model, we create a system capable of learning shared thermal dynamics across heterogeneous inputs. The architecture incorporates skip connections to enhance stability and prediction accuracy. Our approach reduces the need for large simulation datasets while achieving superior performance compared to models trained from scratch. Experimental validation on two case studies material variability and geometric diversity demonstrates significant improvements in generalization, establishing a scalable ML-based solution for intelligent thermal control in manufacturing environments. Moreover, the approach highlights how data-efficient generalization strategies can extend to other industrial applications involving complex physical modeling with limited data.
\end{abstract}


\maketitle

\section{Introduction}

Polyethylene-terephthalate (PET) preforms are small, injection-molded plastic parts that are used to form bottles and containers through a blow molding process \cite{wawrzyniak2020}. Before injection molding, the PET preforms need to be heated to a specific temperature to make them easier to mold. Traditionally, infrared (IR) heating has been the industry standard, but its limitations including energy inefficiency and lack of precise spatial control—have spurred interest in alternative technologies. 

Microwave (MW) heating has emerged as a promising alternative due to its volumetric heating capabilities, faster processing times, and potential for selective energy deposition \cite{yang2014}. Preheating PET preforms carefully and consistently is important for producing high-quality containers before the blow molding process. It provides advantages when the heating is distributed equally along the preform, resulting in better and higher quality bottles by ensuring that the material is uniformly heated and has consistent properties throughout.

This leads to more precise and predictable molding of the preform, resulting in bottles with consistent wall thickness and better clarity. In contrast, uneven heating can lead to defects such as variations in wall thickness, haze, or stress marks, which can compromise the quality and performance of the final product \cite{luo2024}. However, this becomes more challenging to achieve due to the huge variations available in preforms to suit different bottle and container sizes and shapes. Manufacturers can produce preforms in a range of weights and lengths, with different neck finishes and thread designs to match the requirements of various bottling applications \cite{monteix2001}. The specific design of a PET preform will depend on the final container shape and size needed and will be determined by the requirements of the customer.

Recent advancements in deep learning, offer new ways for modeling complex physical phenomena like microwave heating. Yet, training deep neural networks typically requires large datasets, which can be prohibitively expensive and time-consuming to generate through high-fidelity simulations or experiments. 

This work proposes a data-efficient, generalizable deep learning framework for predicting the 2D temperature distribution within PET preforms subjected to microwave heating. Although 3D modeling is theoretically more comprehensive, this study focuses on 2D temperature distribution due to the rotational symmetry of PET preforms during the heating process. Expanding to a full 3D model would not yield significantly different results but would substantially increase computational cost and data requirements. Therefore, a 2D approach offers a more efficient and equally accurate alternative for this specific application.

Our method incorporates two key innovations: (1) transfer learning through fine-tuning, which enables leveraging knowledge from one set of material or geometric conditions to another, and (2) model fusion, where multiple specialized models are combined into a single, robust predictor that generalizes well across unseen scenarios. Two practical case studies are examined:

\begin{itemize}
  \item \textbf{Case Study 1:} Generalization across variations in the heat capacity of PET, relevant for incorporating recycled materials.
  \item \textbf{Case Study 2:} Generalization across different preform geometries, a common variability in manufacturing lines.
\end{itemize}

By using limited datasets (just 450 to 550 samples per category), we demonstrate that our approach significantly reduces data requirements while maintaining high prediction accuracy. The proposed methodology offers a scalable and intelligent alternative to traditional modeling, paving the way for smart, adaptive thermal control systems in plastic manufacturing.

\begin{figure}[t!]
  \centering
  \begin{subfigure}[t]{0.48\linewidth}
    \centering
    \includegraphics[width=0.8\linewidth]{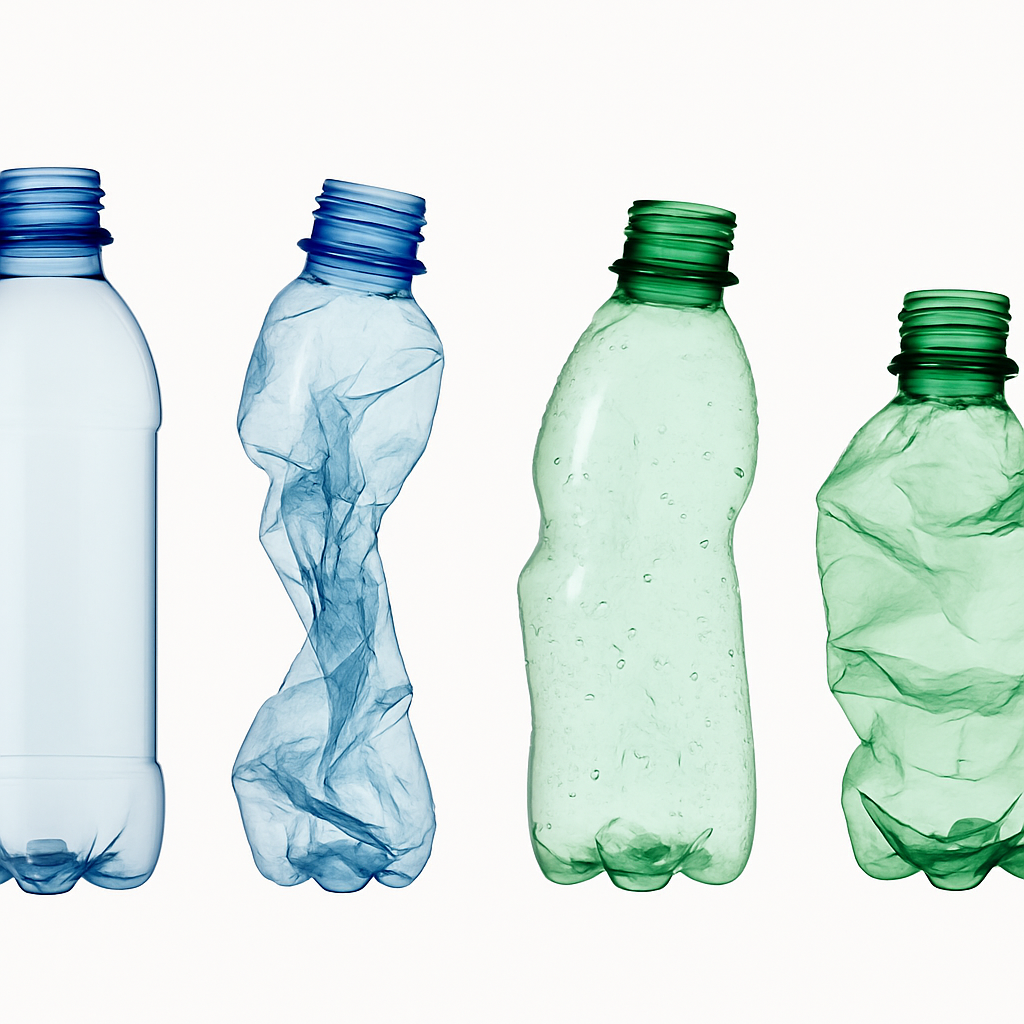}
    \caption*{(a)}
  \end{subfigure}
  \hfill
  \begin{subfigure}[t]{0.48\linewidth}
    \centering
    \includegraphics[width=0.8\linewidth]{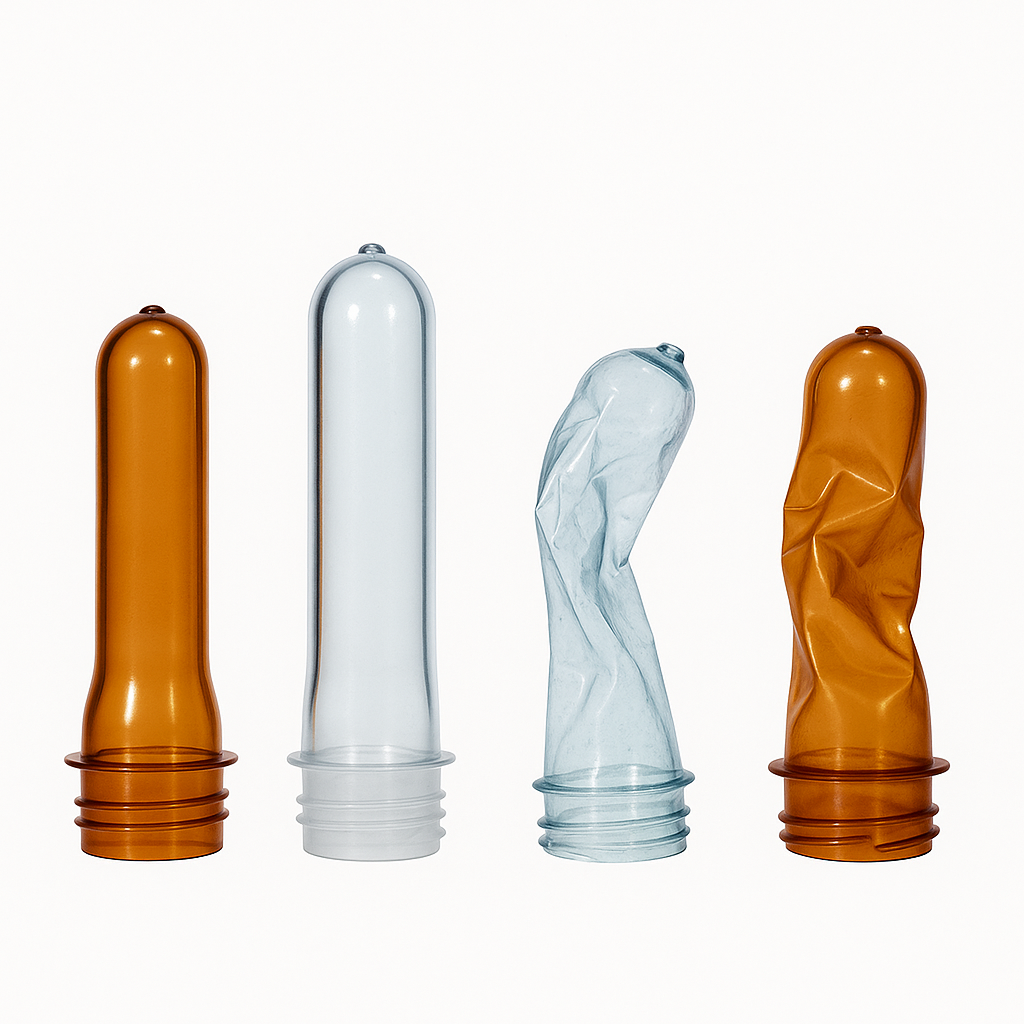}
    \caption*{(b)}
  \end{subfigure}
  \caption{(a) Molded and defective PET bottles; (b) Preforms with different designs.}
  \Description{Side-by-side images showing (a) properly molded and defective PET bottles caused by uneven heating and (b) various PET preforms with different lengths and neck designs.}
  \label{fig:PETBottles}
\end{figure}

\section*{Related Work}

Recent advancements in transfer learning, fine-tuning, and model fusion have significantly enhanced neural network performance across various domains. However, their application to PET preform heating remains limited.

Most existing studies focus on image and signal classification tasks. Liu et al.~\cite{liu2022} and Zhou et al.~\cite{zhou2017} used transfer learning in contexts like garbage sorting and medical imaging. Ghazi et al.~\cite{ghazi2017} and Chakraborty et al.~\cite{chakraborty2021} demonstrated adaptability in plant identification and human action recognition, while Korzh et al.~\cite{korzh2018} and Whitney et al.~\cite{whitney2019} showed improved performance using ensemble models.

Emerging methods such as model merging~\cite{geyer2019} and AdapterFusion~\cite{pfeiffer2020} further improve task generalization. Ge and Yu~\cite{ge2017} and Zhai et al.~\cite{zhai2020beauty} explored multi-fidelity and multi-channel fusion to enhance predictive accuracy.

Machine learning use in industrial heating is still rare. Notable efforts include Hsieh~\cite{hsieh2023}, who applied deep reinforcement learning to control blow molding temperatures, and Zhai et al.~\cite{zhai2020temp}, who used transfer learning in heating furnace prediction. Di Barba et al.~\cite{dibarda2023} also introduced neural metamodels for adaptive induction heating control.

Traditionally, industrial heating systems have relied on physics-based models and heuristic control strategies. However, these often struggle with dynamic production environments. In PET blow molding specifically, infrared (IR) heating remains the dominant method for preheating preforms. Conventional infrared (IR) ovens are commonly used to heat PET preforms prior to blow molding. However, IR heating is limited to surface absorption, offers slow thermal response, and often struggles with achieving uniform radial temperature profiles~\cite{yang2014}. In contrast, microwave (MW) heating penetrates deeply into the material, enabling volumetric absorption and significantly shorter heat-up times—up to 80 ~\cite{garcia2022}.

\section{Methodology}
\subsection{Applicator Design and Simulation}

The design and functionality of the applicator play a critical role in achieving precise heating patterns, particularly in applications requiring uniform temperature distribution. The proposed applicator consists of a rectangular cavity with internal dimensions of 250\,mm in width, 190\,mm in length, and 150\,mm in height. It is equipped with dielectric slabs that fine-tune the electromagnetic field distribution.

The cavity is energized via a Type-N coaxial antenna, located at the center of the bottom wall and aligned along the \textit{z}-axis. It is configured to generate the TE\textsubscript{101} electromagnetic mode at 915\,MHz~\cite{garcia2022}, a standard frequency in industrial microwave applications.

The heating process is highly dependent on the geometry of the PET preform, including parameters such as wall thickness, neck dimensions, and overall shape of which directly affects the characteristics of the final container. To manipulate the field distribution, the applicator includes dielectric slabs composed of two stacks of 16 PTFE (polytetrafluoroethylene) sheets. These slabs are microwave-transparent and are oriented parallel to the \textit{y}-axis.

The slabs are placed symmetrically on either side of the preform, with their positions adjustable along the \textit{x}-axis to control the distance from the preform. Each slab measures 25\,mm in width, 190\,mm in length, and 5\,mm in height.

The complete system was modeled and simulated using Ansys HFSS, a high-frequency electromagnetic simulation platform, to optimize design parameters and validate heating effectiveness. A key innovation in the system is the use of dielectric slabs as near-field focusing lenses~\cite{bakerjarvis2012}, which allow for precise manipulation of electromagnetic waves through reflection, refraction, and diffraction~\cite{garcia2022}.

\begin{figure}[t]
  \centering
  \includegraphics[width=1\linewidth]{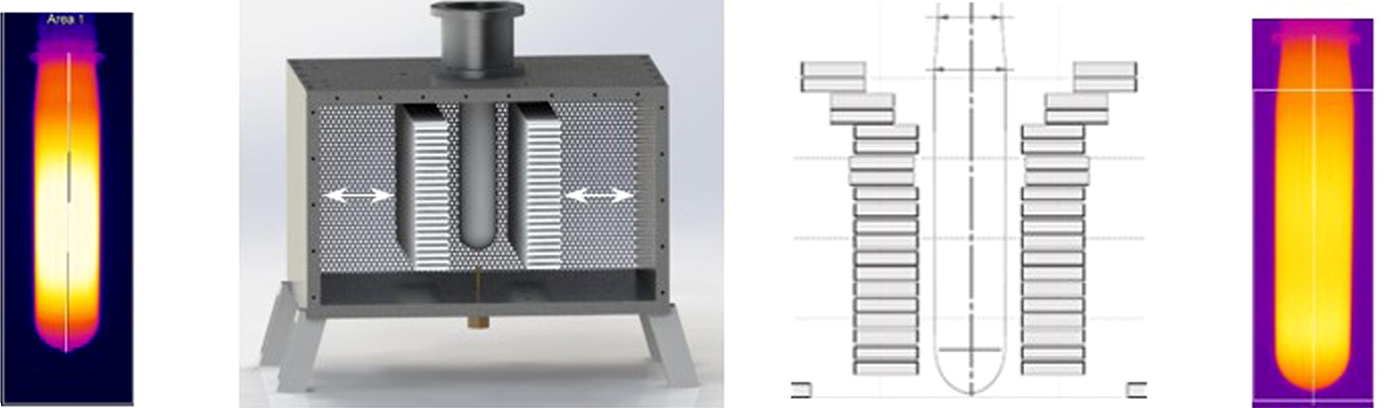}
  \caption{Schematic representation of the microwave heating optimization process for PET preforms. The default temperature profile (left) shows uneven heating typically observed without slabs modifications. A custom-designed microwave cavity with adjustable dielectric slabs (center-left) enables fine-tuning of the electromagnetic field distribution. The optimized slab con-figuration (center-right) is tailored to the preform geometry, leading to a significantly im-proved and more uniform temperature profile (right). . Adapted from García-Baños et al.~\cite{garcia2022}.}
  \Description{Schematic showing four stages of the PET preform heating process: (1) default temperature profile, (2) a microwave cavity with dielectric slabs, (3) optimized slab configuration, and (4) improved uniform temperature profile.}
  \label{fig:microwave}
\end{figure}

\subsection{Generalization and Model Fusion Methodology}

In machine learning, the challenge of model generalization extends beyond simply performing well on unseen data. A regression model that generalizes effectively can extrapolate to new, related datasets. A more advanced challenge involves merging multiple locally generalized models into a single, globally generalized model—a process known as model fusion. By combining diverse models, the overall predictive performance can be enhanced, as each model contributes unique strengths to the final output.

Several model fusion techniques exist~\cite{huang2009}, including:

\begin{itemize}
  \item \textbf{Voting:} Typically used in classification tasks, this method aggregates predictions from multiple models and selects the majority outcome.
  \item \textbf{Averaging:} Applicable to both regression and classification, it smooths predictions and helps reduce overfitting.
  \item \textbf{Stacking:} This method involves training a second-level model on the outputs (predictions) of several base models to form a meta-learner.
\end{itemize}

We selected stacking over simple voting or averaging ensembles because stacking trains a meta‑learner to combine the outputs of base models in a non‑linear fashion, often outperforming fixed combination rules—especially in regression contexts~\cite{huang2009}. Voting (or averaging) only computes the mean or majority output and cannot learn how to weight or combine predictions in task-specific ways.

\subsection{Data Collection and Predictor Model Training}

Our methodology for training and merging pretrained predictor models begins with selecting three distinct variations of the target variable, representing low, medium, and high values. Initial data collection is conducted using Design of Experiments (DOE) to reduce the dimensionality of the input space, particularly focusing on the slab positions. Among various DOE strategies, we employed Latin Hypercube Sampling (LHS)~\cite{huntington1998}, which is known for effectively covering large parameter spaces with minimal experimental runs. 
As previously described, the microwave heating system was simulated using Ansys HFSS to generate the training data. The input features to the predictor model include:

\begin{itemize}
  \item Slab positions along the \textit{x}-axis (continuous variables),
  \item Preform geometrical attributes such as length, weight, and neck dimensions (when applicable),
  \item Material-specific properties such as heat capacity (used in Case Study 1).
\end{itemize}

\par\noindent\textbf{Note:} We utilized a 2D axisymmetric simulation to reduce computational complexity, while still capturing the full axial thermal behavior of PET preforms. Due to the nearly rotationally symmetric geometry, a full 3D simulation would yield essentially the same temperature results as the 2D model, but at much higher cost.

In both case studies, the output (target) is the spatial temperature field, represented as a set of continuous temperature values (in °C) at 32 discrete surface points along the PET preform. This setup defines a regression task in which each model predicts fine-grained temperature distributions based on configuration inputs.

For each case, an initial model was trained using data from the first variation and validated on a separate unseen dataset to verify accuracy. The same model architecture and pretrained weights were then fine-tuned for the remaining two variations, ensuring consistency across training phases.

Model performance was evaluated using standard regression metrics, including:
\begin{itemize}
  \item Root Mean Squared Error (RMSE),
  \item Mean Absolute Error (MAE),
  \item Coefficient of Determination ($R^2$).
\end{itemize}

\begin{figure}[t]
  \centering
  \includegraphics[width=1\linewidth, trim=40 120 100 60, clip]{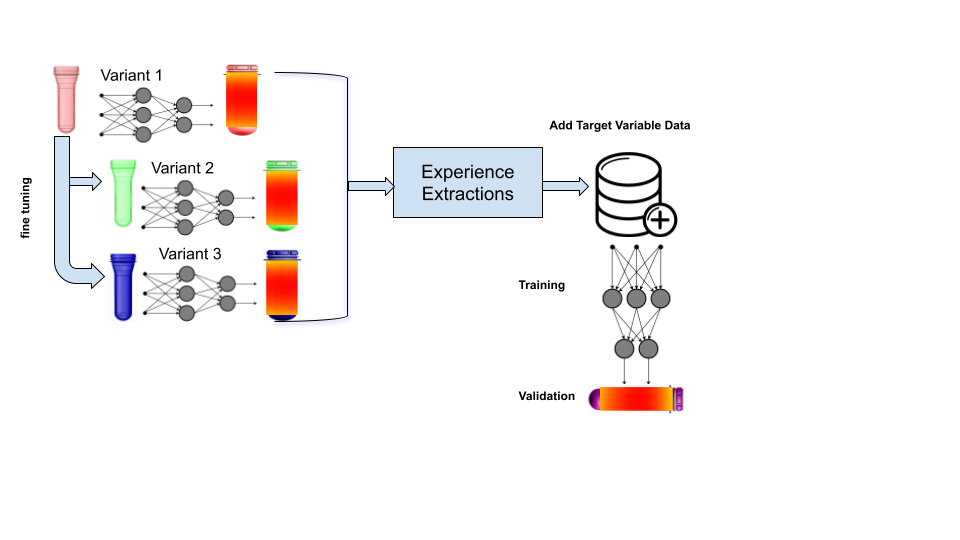}
  \caption{Model fusion workflow combining outputs from variant-specific models into a unified generalized predictor.}
  \Description{Diagram illustrating model fusion steps: separate predictor models generate outputs, which are scaled and merged with design data to train a final generalized model.}
  \label{fig:fusion}
\end{figure}

\subsection{Fusion Implementation and Neural Network Architecture}

Figure~\ref{fig:fusion} illustrates the process of model fusion, beginning with the individual training of predictor models for each variation. The subsequent step involves \textit{experience extraction}, wherein a new Design of Experiments (DOE) setup is constructed and each trained model is tasked with making predictions based on this fresh experimental design.

These predicted outputs are appropriately scaled, and corresponding target variables (associated with each preform variation) are integrated into the dataset. This fusion process merges information extracted from each pretrained model and incorporates variation-specific characteristics to build a more generalizable predictor. The goal is to extend the size and diversity of the initial dataset by synthesizing new samples using model predictions, effectively augmenting the training data.

A new, generalized predictor model is then trained on this merged dataset. By leveraging the fused information, the model gains a broader understanding of the input–output relationships and demonstrates improved prediction accuracy and generalization performance across unseen configurations.

To evaluate the effectiveness of this fusion-based learning strategy, the final model is tested on a preform variation that was not included in any prior training phase. In this context, ``experience'' refers to the predictive knowledge embedded within each pre-trained model. This knowledge is utilized to simulate outcomes for new scenarios, which are then compiled into an expanded training dataset. The outputs are rescaled and coupled with their corresponding target labels, forming a comprehensive dataset enriched with structural and material variability.

Input parameters used in all predictors—including slab positions, preform geometries, and material properties—were obtained from Ansys HFSS simulations and verified against known manufacturer specifications. All data generation and transformation steps were automated using scripting and batch processing to ensure reproducibility and scalability.

\begin{figure}[t!]
  \centering
  \begin{subfigure}[t]{\linewidth}
    \centering
    \includegraphics[width=0.8\linewidth, height=1.8cm]{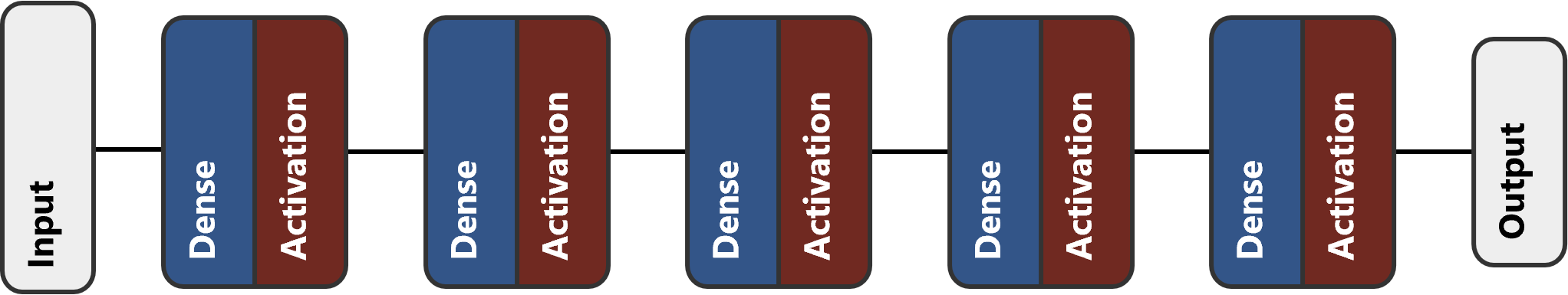}
    \caption*{(a)}
  \end{subfigure}
  \vspace{0.5em}
  \begin{subfigure}[t]{\linewidth}
    \centering
    \includegraphics[width=0.8\linewidth, height=2.2cm]{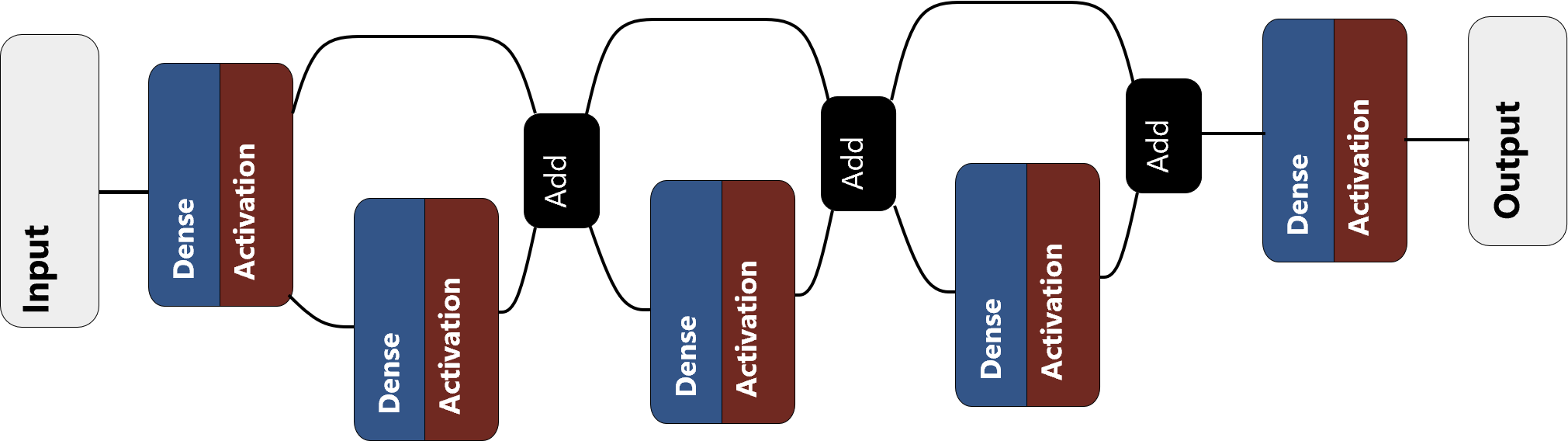}
    \caption*{(b)}
  \end{subfigure}
  \caption{Comparison of MLP architectures: (a) standard and (b) with skip connections ("Add") for improved gradient flow and stability.}
  \Description{Two neural network diagrams: (a) a basic multilayer perceptron and (b) a similar architecture with skip connections shown by “Add” nodes.}
  \label{fig:mlpSkip}
\end{figure}

To develop the predictor model, we compared two neural network architectures, shown in Figures~\ref{fig:mlpSkip}a and~\ref{fig:mlpSkip}b, to determine the more efficient option. Both networks shared identical input and output configurations, the same number of hidden layers, and identical activation functions.

The first architecture is a standard Multilayer Perceptron (MLP), while the second incorporates \textit{skip connections}. These connections, also known as \textit{residual connections}, are designed to address challenges such as the vanishing gradient problem. They enable more efficient learning by allowing information to bypass intermediate layers and directly propagate forward~\cite{he2015}.

Table~\ref{tab:evaluation_metrics} presents the evaluation metrics—Root Mean Squared Error (RMSE), Mean Absolute Error (MAE), and the coefficient of determination ($R^2$)—on the test dataset. The results show that the MLP with Skip Connection consistently outperforms the standard MLP, achieving lower prediction errors and higher predictive accuracy across all metrics.

\begin{table*}
  \caption{Evaluation metrics for standard MLP and MLP with Skip Connection models}
  \label{tab:evaluation_metrics}
  \begin{tabular}{lccc}
    \toprule
    Metric & Standard MLP & MLP with Skip Connection & Improvement \\
    \midrule
    RMSE   & 0.185 & 0.052 & $\downarrow$ 72\% \\
    MAE    & 0.148 & 0.039 & $\downarrow$ 74\% \\
    $R^2$  & 0.91  & 0.98  & $\uparrow$ 7.7\% \\
    \bottomrule
  \end{tabular}
\end{table*}

\section{Case Studies for Model Validation}

To evaluate the robustness and generalization capability of the proposed approach, we conducted two case studies: one focusing on variations in material characteristics specifically, heat capacity differences between virgin and recycled PET—and the other on geometrical variations in PET preforms. These case studies demonstrate how the model adapts to both material-related and shape-related differences commonly encountered in production environments.

\subsection{Case Study 1: Material Characteristics of PET}

This case examines model generalization with respect to PET material variations. While virgin PET is standard in preform production, environmental concerns have increased the use of recycled PET (rPET). Although rPET aims to mimic virgin PET properties, the recycling process can introduce impurities, poor sorting, and thermal degradation~\cite{venkatachalam2012}, leading to structural inconsistencies.

One key property affected is heat capacity—the ability to absorb and retain heat. Virgin PET typically has higher heat capacity due to fewer structural defects, while rPET often shows reduced values due to crystallization disruption and contamination.

\begin{figure}[t!]
  \centering
  \begin{subfigure}[t]{0.5\linewidth}
    \centering
    \includegraphics[width=\linewidth]{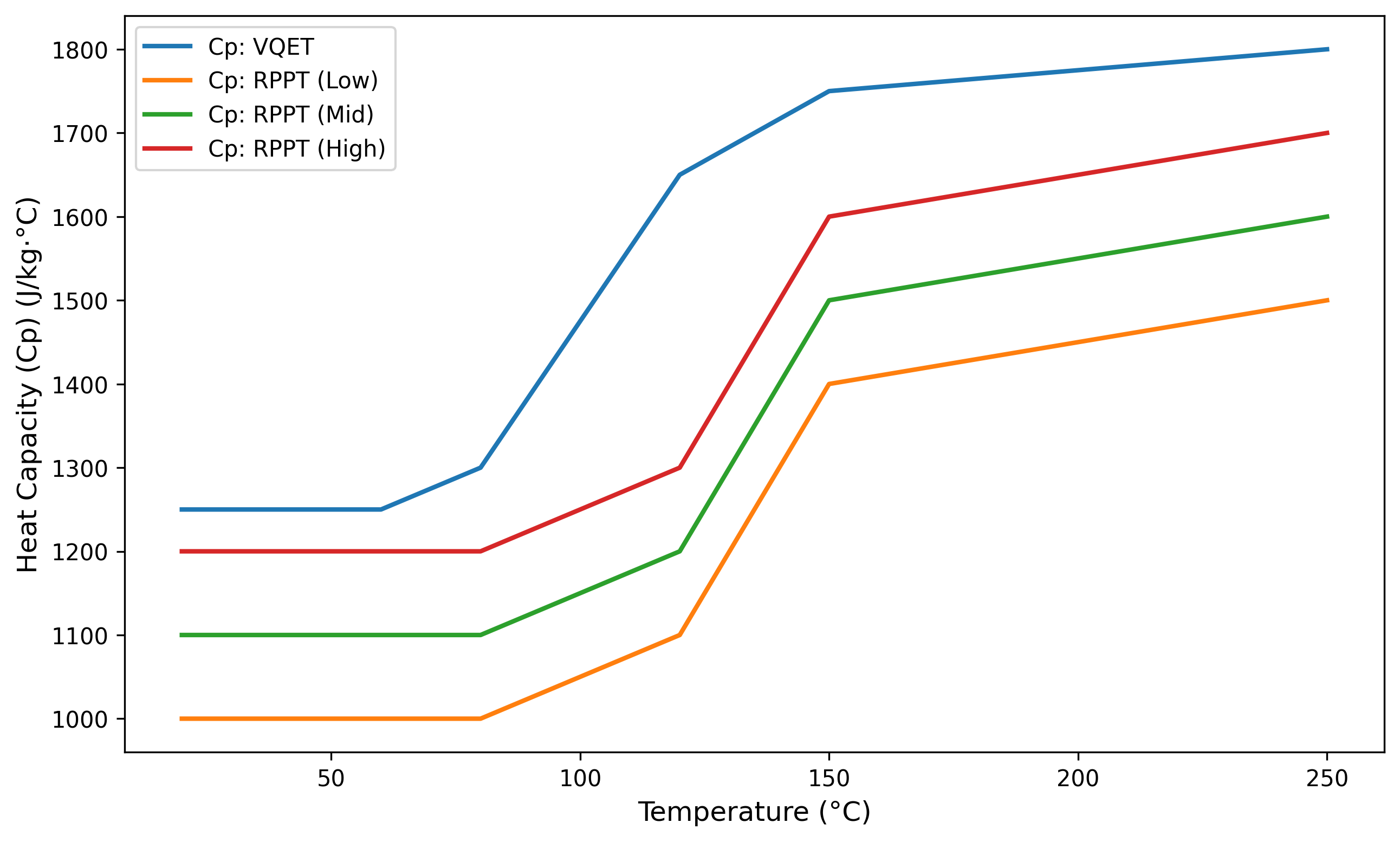}
    \caption*{(a)}
  \end{subfigure}
  \hfill
  \begin{subfigure}[t]{0.46\linewidth}
    \centering
    \includegraphics[width=\linewidth]{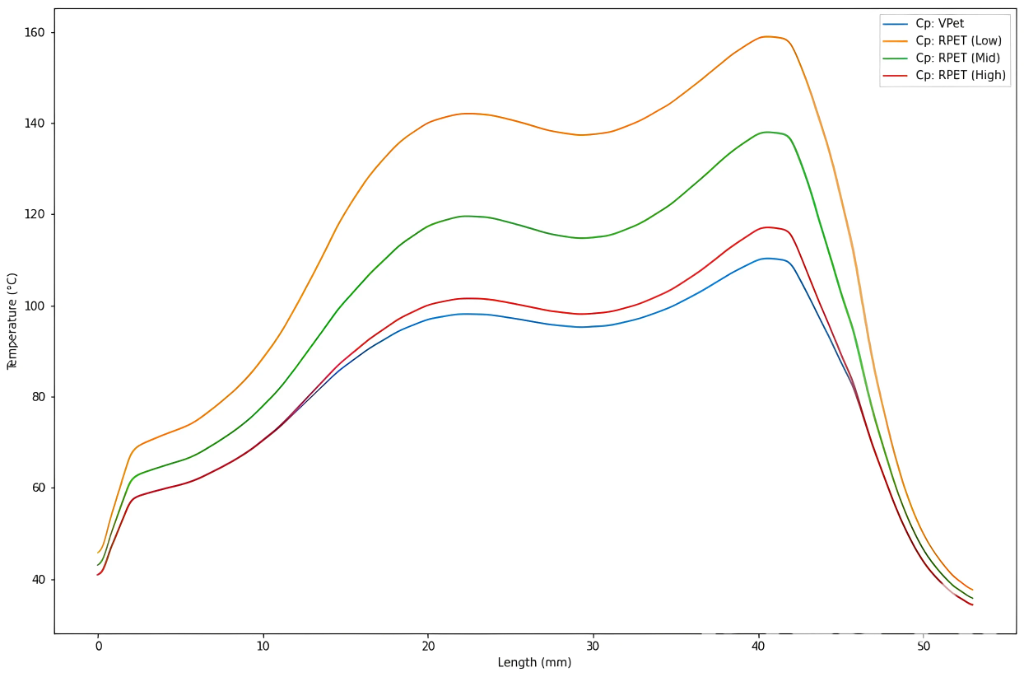}
    \caption*{(b)}
  \end{subfigure}
  \caption{(a) Heat capacity vs. temperature for virgin and recycled PET. (b) Resulting temperature profiles along preform length under different heat capacity assumptions.}
  \Description{Two subfigures: (a) line plot of heat capacity vs. temperature for virgin and recycled PET; (b) corresponding preform temperature profiles under different heat capacity settings.}
  \label{fig:heatCapacityEffect}
\end{figure}

Since rPET data are limited, we modeled three plausible heat capacity variations as temperature-dependent functions and compared them with a reference virgin PET curve. These models, shown in Figure~\ref{fig:heatCapacityEffect}, reveal that variations in heat capacity significantly impact thermal distribution, influencing heating model accuracy.

\subsection{Case Study 2: Geometrical Variations of Preforms}

This case investigates the model’s generalization to varying preform geometries—critical in industrial settings where preforms differ in size, weight, and design.

Four representative geometries were selected, varying in length, wall thickness, weight, and curvature. These were used to test the generalized model’s robustness, particularly its ability to handle unseen shapes without retraining. The results demonstrate the model’s flexibility and accuracy across diverse preform designs.

\begin{table*}
  \caption{Heat capacity and temperature array definitions for each dataset category used in Case Study 1}
  \label{tab:heatcapacity}
  \begin{tabular}{lccc}
    \toprule
    Heat Capacity Category & Heat Capacity Array [J/kg°C] & Temperature Array [°C] & Dataset Size [\# of signals] \\
    \midrule
    Low Cp  & [1000, 1050, 1100, 1350, 1450] & [80, 100, 120, 150, 250] & 550 \\
    Mid Cp  & [1100, 1150, 1200, 1500, 1600] & [80, 100, 120, 150, 250] & 450 \\
    High Cp & [1250, 1300, 1650, 1750, 1800] & [80, 100, 120, 150, 250] & 450 \\
    \bottomrule
  \end{tabular}
\end{table*}
\subsection{Adaptation to Material and Geometry-Specific Characteristics}

This section describes how the proposed approach was validated and fine-tuned for the material and geometrical differences introduced in the previous case studies.

All training and validation datasets used in this study were generated entirely through high-fidelity electromagnetic and thermal simulations in Ansys HFSS. Each simulation run was automated using parametric scripting, enabling efficient, reproducible data generation at scale. This simulation-driven approach allowed us to compile datasets of over 6,000 labeled samples while avoiding the time and cost of physical experiments.

\paragraph{Case Study 1: Adaptation to Material Characteristics.}

The first case addresses adaptation to variations in PET heat capacity. A base predictor model was initially trained using a mid-range heat capacity dataset comprising 550 samples. The model was subsequently fine-tuned using 450 additional samples from datasets representing low and high heat capacity conditions, respectively.

Each of these three models was trained using different slab position settings to predict temperature values at 32 predefined spatial locations along the PET preform surface.

To assess generalization performance, each model was evaluated on an unseen test dataset. The heat capacity values used to define the low, medium, and high material categories are summarized in Table~\ref{tab:heatcapacity}.

After confirming high accuracy across the locally trained models, a new Design of Experiments (DOE) was constructed, generating 2{,}000 synthetic data points for each model. Each model was then used to predict outcomes on this DOE, resulting in a merged dataset of 6{,}000 samples. This unified dataset included predicted temperature distributions, associated heat capacity values, and relevant input features such as slab positions.

\begin{figure}[t]
  \centering
  \includegraphics[width=\linewidth]{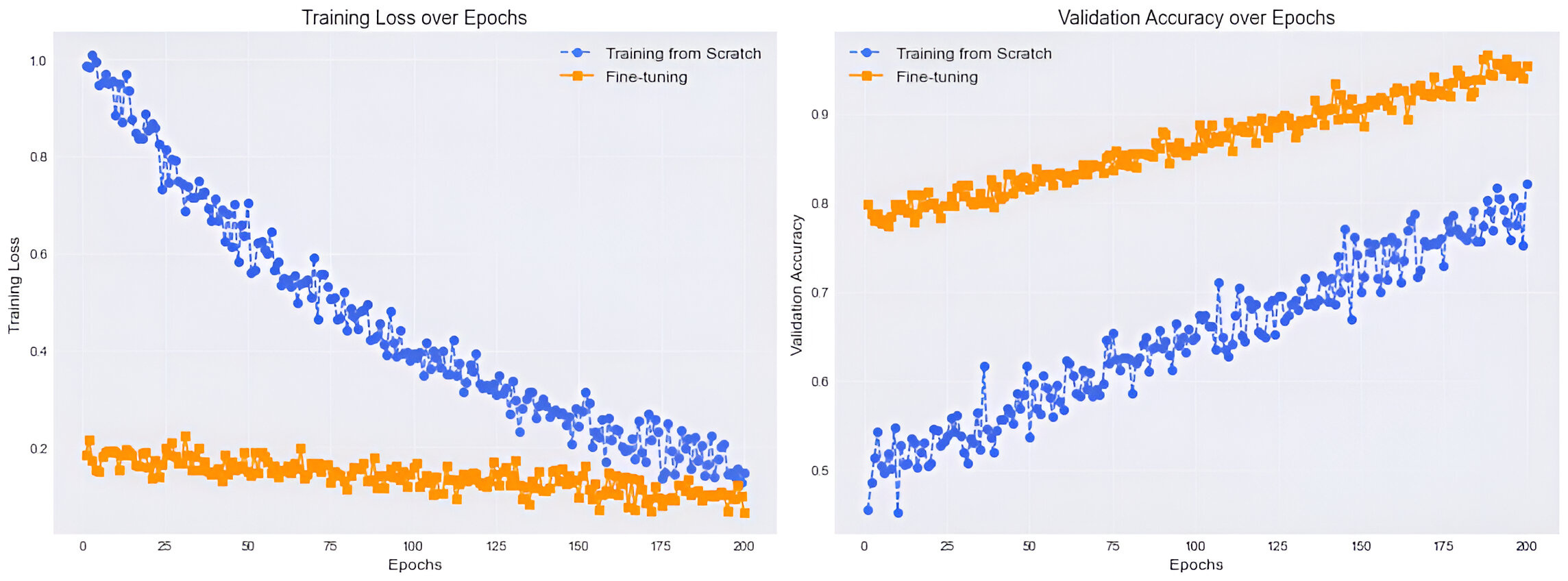}
  \caption{Training Loss and validation Accuracy for Case 1}
  \Description{Line plot showing training loss decreasing and validation accuracy increasing over epochs. The global model outperforms the baseline, indicating better generalization to unseen heat capacity profiles.}
  \label{fig:globalModel}
\end{figure}

A global predictor model was subsequently trained using this enriched dataset, as illustrated in Figure~\ref{fig:globalModel}. For benchmarking purposes, the global model was compared to a baseline model trained from scratch using a combined real dataset of 1{,}950 samples—comprising 625 samples from the low, 700 from the mid, and 625 from the high heat capacity categories.

To validate performance, the models were tested on a new, previously unseen heat capacity profile not included in any training phase.

\begin{figure}[t]
  \centering
  \includegraphics[width=0.8\linewidth]{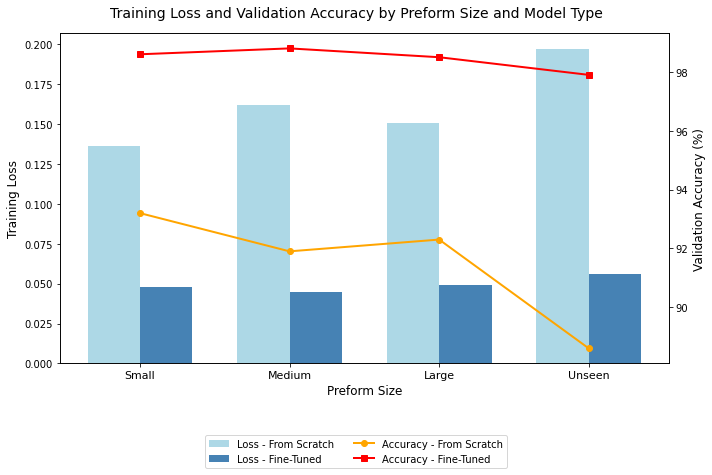}
  \caption{Training Loss and validation Accuracy for Case 2}
  \Description{Graph showing training loss and validation accuracy over epochs for Case 2. The global model achieves stable training performance and maintains high accuracy when tested on a new preform geometry not seen during training.}
  \label{fig:evaluation}
\end{figure}

\paragraph{Case Study 2: Adaptation to Geometrical Variations.}

In the second case study, we evaluated the generalization capabilities of the model with respect to PET preform geometry. Three preform sizes—small, medium, and large—were used for model training and fine-tuning.

Following the same DOE-based experience extraction process described in Case Study 1, a new synthetic dataset comprising 6{,}000 samples was compiled. Each sample included inputs such as slab positions and critical geometrical attributes (e.g., weight, neck length, and wall thickness).

A global model was then trained on this dataset to generalize predictions across a wide range of preform geometries. For validation, the model was tested on a preform geometry not present in any of the training datasets, thereby assessing its ability to extrapolate across shape-based variability as shown in Figure~\ref{fig:evaluation}.

\section{Conclusions and Future Work}
This paper presented a data-efficient generalization technique for regression tasks, applied to temperature prediction in microwave preheating of PET preforms before blow molding. By combining fine-tuning and model fusion, the approach achieved accurate predictions across diverse material and geometrical variations using significantly fewer samples than traditional methods. Results confirm strong generalization to unseen variants, offering a scalable solution for data-limited industrial applications.
This method through integration of transfer learning and model fusion is well-suited for physical modeling tasks where simulation data are costly or system variability is high.
Future work will focus on handling dynamic material and environmental variations, exploring adaptive, real-time updates, and advancing fusion strategies. A key limitation is the need to fine-tune multiple models; we aim to develop architectures capable of generalizing across variants using a single training pass on a unified dataset, improving scalability.
While the dataset is proprietary, it was generated via Ansys HFSS simulations using Latin Hypercube Sampling. Each variant-specific dataset (450–550 samples) was fine-tuned independently, then merged via model outputs for fusion. Models were built in TensorFlow, trained on a workstation with an RTX 3080 GPU. Although the dataset cannot be shared, pseudo-code and synthetic data will be released in the future. Interested researchers may contact the corresponding author via email for further information.

\subsection*{Acknowledgments}
The authors wish to thank Thomas Albrecht and Guenter Winkler for their support, fruitful discussions, and useful advices.

\subsection*{Disclosure of Interests}
The first author is pursuing a PhD at Deggendorf Institute of Technology in collaboration with Krones AG, which funded the research presented in this article. The second author, a faculty member at Deggendorf Institute of Technology, contributed in an academic supervisory capacity. The authors declare that they have no other competing interests.

\section*{Declaration on Generative AI}
 During the preparation of this work, the author(s) used ChatGPT (GPT-4) in order to: assist with language editing and LaTeX formatting. After using this tool, the author(s) reviewed and edited the content as needed and take(s) full responsibility for the publication’s content.

\bibliography{sample-ceur}

\end{document}